\documentclass[a4paper,11pt]{article}

\usepackage[english]{babel}
\usepackage[a4paper]{geometry}
\usepackage{amsmath, amsfonts, systeme, mathtools}
\usepackage{setspace}
\usepackage{graphicx}
\usepackage[colorlinks=true, citecolor=blue, linkcolor=blue]{hyperref}
\usepackage[biblabel]{cite}

\usepackage{algorithm, algorithmic}
\usepackage{enumitem}
\usepackage{listings}
\usepackage{xcolor}
\usepackage{sectsty}
\usepackage{notoccite}
\usepackage{indentfirst}

\usepackage{libertine}

\usepackage{titlesec}
\titlespacing{\section}{0pt}{0pt}{0pt}
\usepackage{caption}
\usepackage{subcaption}
\begin{document}
	
	{\Large \textsf{Image-Based Structural Analysis Using Computer Vision and LLMs: \textit{\\PhotoBeamSolver}}}\\[1em]
	{\small \textsf{ALTAMIRANO-MUÑIZ EMILIO FERNANDO}}\\
	{\small \textsf{mail: amz [at] gmx [dot] cn}}\\

\vspace{1em}
\noindent
\textbf{Abstract}

\small
This paper presents the development of a documented program capable of solving idealized beam models, such as those commonly used in textbooks and academic exercises, from drawings made by a person. The system is based on computer vision and statistical learning techniques for the detection and visual interpretation of structural elements.\\
Likewise, the main challenges and limitations associated with the integration of computer vision into structural analysis are analyzed, as well as the requirements necessary for its reliable application in the field of civil engineering.\\
In this context, the implementation of the \textit{PhotoBeamSolver} program is explored, and the current state of computer vision in civil engineering is discussed, particularly in relation to structural analysis, infrastructure inspection, and engineering decision-support systems.

\normalsize

\section{Introduction}
\label{cap:1}
Computer vision and artificial intelligence (AI) is a rapidly growing field with the potential to change the way civil engineering projects are designed, built, and maintained. The use of computer vision techniques enables the automation of image and video analysis and interpretation, which can increase the efficiency and accuracy of tasks such as surveying, structural analysis, anomaly detection, etc. These forms of computer vision are empowered by artificial intelligence, i.e., machine learning/deep learning.\\
In the last two decades, there have been overwhelming advances in areas of computer vision, artificial intelligence, and this has eventually affected civil engineering. This translates into a variety of tools for different fields of this discipline. It can go as far as structural safety monitoring \cite{NIAN2021103838, cv_struct} to sophistication in geographic information systems (GIS).\\
The state of the art in computer vision and artificial intelligence in civil engineering involves the use of deep learning models to analyze and interpret images and videos of infrastructure on construction sites. This includes tasks such as object detection and recognition, semantic segmentation, and depth estimation.\\
One application of computer vision in the field is the detection and measurement of cracks, deformations, corrosion \cite{corrosion_detec}, and other damage in bridges or their components \cite{DBLP:journals/corr/abs-1806-06820}, as well as in roads and buildings \cite{SPENCER2019199}. This can be done using convolutional neural networks (CNN's) trained to detect and classify different types of damage.\\
Another application is in site management with activities such as: progress monitoring \cite{quick_bitacora}, identification of potential hazards or safety violations \cite{rempro}, and automation of worker and equipment tracking. This can be achieved using techniques like semantic segmentation to identify different objects in the scene, and object tracking to follow the movement of workers over time.\\
Computer vision algorithms have also been used to inspect and monitor large-scale infrastructure projects such as bridges, dams, and skyscrapers. The techniques used for the activities mentioned are 3D reconstruction, which can be used to create 3D models of the structure, and change detection to identify and track changes over time.\\
In the field of surveying and cartography, computer vision is used to extract aerial and satellite imagery for activities like digital surface modeling (DSM's), detection of constructions and infrastructure, etc.\\
Several applications of deep learning and machine learning involve techniques that merge with computer vision, although there are more, to improve safety in construction, efficiency, and decision-making on-site.\\
\begin{figure}[!t]
	\minipage{0.32\linewidth}
	\includegraphics[width=\linewidth]{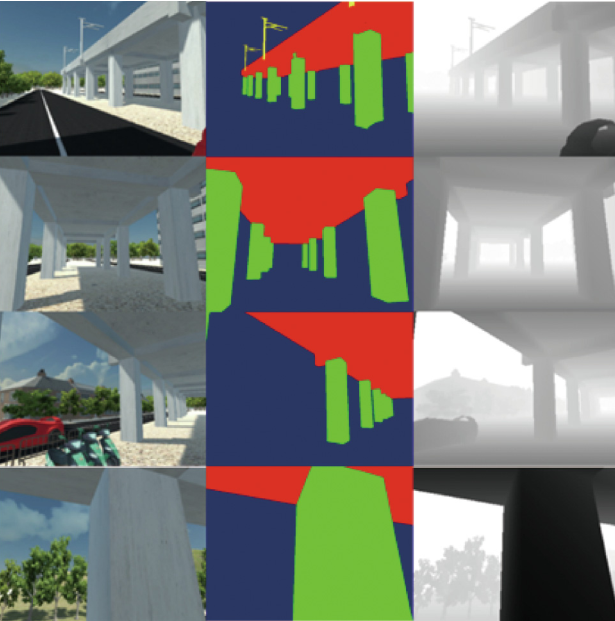}
	\endminipage\hfill
	\minipage{0.32\linewidth}
	\includegraphics[width=\linewidth]{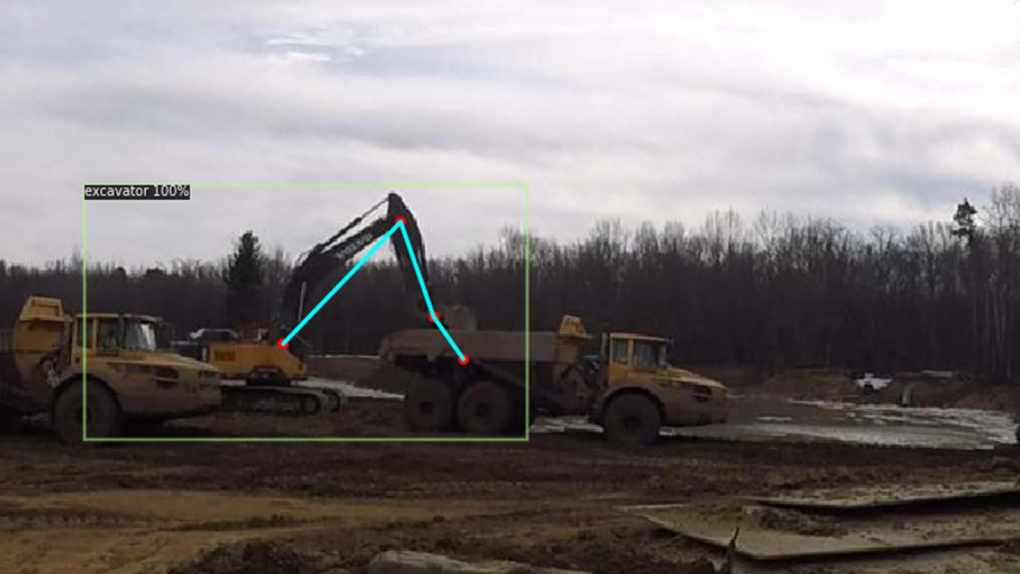}
	\endminipage\hfill
	\minipage{0.32\linewidth}
	\includegraphics[width=\linewidth]{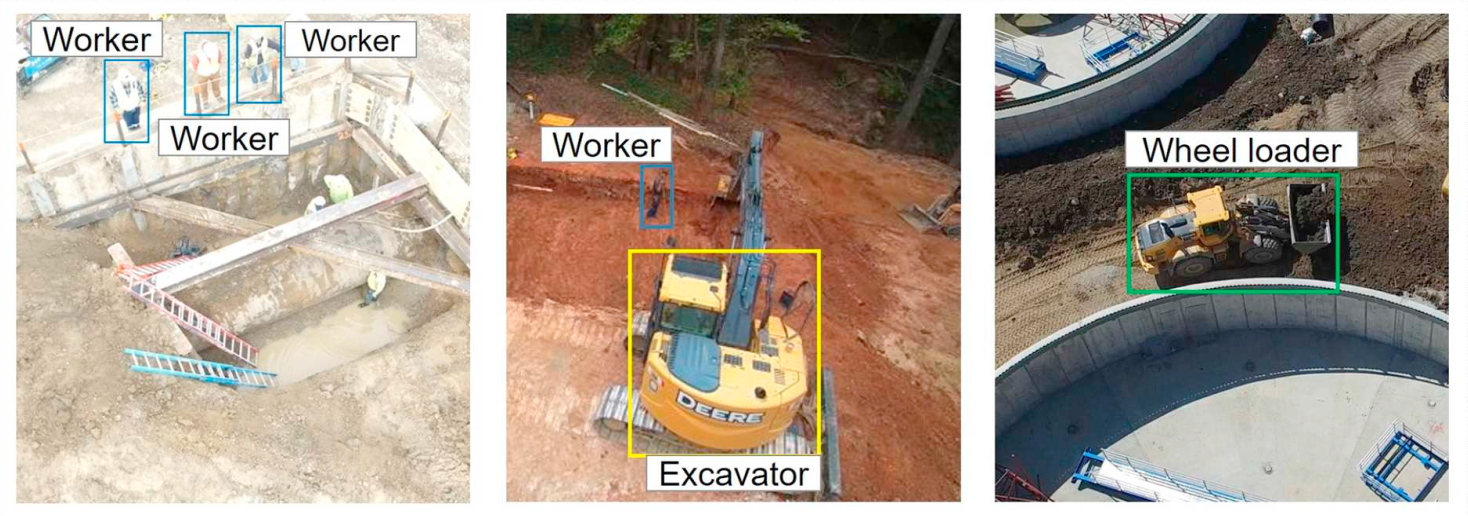}
	\endminipage
	\caption[Ejemplos de la visión de computadoras en la ingeniería civil]{On the left, structural component recognition is applied on a bridge \cite{DBLP:journals/corr/abs-1806-06820}. In the center, the detection of an excavator arm is performed \cite{doi:10.1061/(ASCE)CP.1943-5487.0001060}. On the right, detection of workers and machinery \textit{in situ.} \cite{rempro}}
\end{figure}
Computer vision makes extensive use of machine learning algorithms, which are also addressed in this work. In machine learning, and especially in deep learning, artificial neural networks are used. Different areas of computer vision were mentioned, but outside of that subset there are also many uses of machine learning such as post-seismic damage assessment \cite{dam_asse_seismic} using predictive models. These models are also useful for estimating seismic intensity measures for the same event in different regions \cite{ground_mot_NN}.\\
For several decades, there has been intrigue about how the rise of neural networks could be integrated into structural analysis. Curiosities such as predicting the location of a moment \cite{nn_in_strcu_2}, predicting optimal designs \cite{nn_in_strcu, HAJELA1991473}, or predicting capacities or displacements \cite{nasa_estr} can be mentioned as the \textit{beginnings} of finding a utility for artificial intelligence in structural engineering, eventually reaching larger applications such as the automation of optimal structural designs \cite{sim_struc}. However, the tools that have most widely reached practice are not based on artificial intelligence, but rather are only calculation aids.\\
Regarding the solution of beams and structures, there are multiple tools, some more versatile than others, some more expensive than others, and some easier to use than others. Several free tools exist on the internet such as SkyCiv, MechaniCalc, ClearCalcs, or IndeterminateBeam \cite{Bonanno2021}, the latter being used for the solution in this program. Likewise, software with greater computational power exists, such as SAP2000, STAAD, ANSYS, ETABS, Autodesk Simulation, ADINA, FreeCAD, etc.\
Given the current conditions of artificial intelligence, a question arises: \textbf{how can a structural analysis be performed using only a photograph?} This is where \textit{PhotoBeamSolver} breaks in.\\
Much of the software for engineering use is paid and also expensive; several of the calculators that can be found online, such as those mentioned previously, may be free to use but with significant restrictions.\\
The free tools that do exist may require specific technical knowledge, e.g., knowing a specific programming language different from what a user may know, or operating in an unfriendly user environment, among others.\\
The program \textit{PhotoBeamSolver} can be welcomed by professionals and students who require a quick verification and/or calculation without having to experience the typical inconveniences that other options imply, and it can contribute to the design process.
\section{Formulation}
\label{cap:2}
\textit{PhotoBeamSolver} is a hybrid image-to-analysis pipeline for structural beam analysis. It combines computer vision for diagram interpretation with large language models (LLMs) for symbolic reasoning. The architecture consists of two distinct settings: (a) perception through a custom-trained object detection model, and (b) perception and reasoning through an LLM.\\
The overall workflow converts a raw beam image into a structured mechanical representation, which is subsequently transformed into equilibrium equations and closed-form solutions.
Structural elements are identified using a custom-trained YOLO-based convolutional neural network or an LLM. The detector was trained specifically on annotated beam diagrams containing supports, concentrated loads, distributed loads, and geometric references. Each training sample includes bounding box coordinates and categorical labels corresponding to structural components.\\
The detection model performs single-stage inference, simultaneously predicting bounding box coordinates, class probabilities, and confidence scores. This formulation enables the localization and classification of multiple structural elements within a single forward pass. Redundant detections are removed via non-maximum suppression to ensure a consistent structural interpretation.\\
The choice of a YOLO-based architecture is motivated by its computational efficiency and its suitability for multi object detection in structured diagrams. Unlike region detectors, which require multiple forward passes over region proposals, the selected architecture maintains real-time inference capability while preserving sufficient localization accuracy for structural construction.\\

\begin{figure}[h!]
	\centering
	
	\begin{subfigure}[t]{0.59\linewidth}
		\centering
		\includegraphics[width=\linewidth]{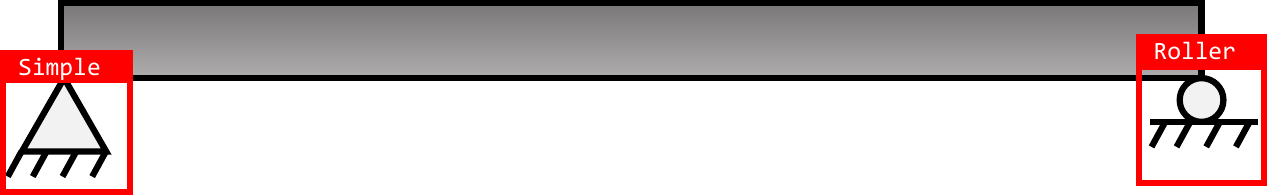}
		\label{objdetec}
	\end{subfigure}
	\hfill
	\begin{subfigure}[t]{0.39\linewidth}
		\centering
		\includegraphics[width=\linewidth]{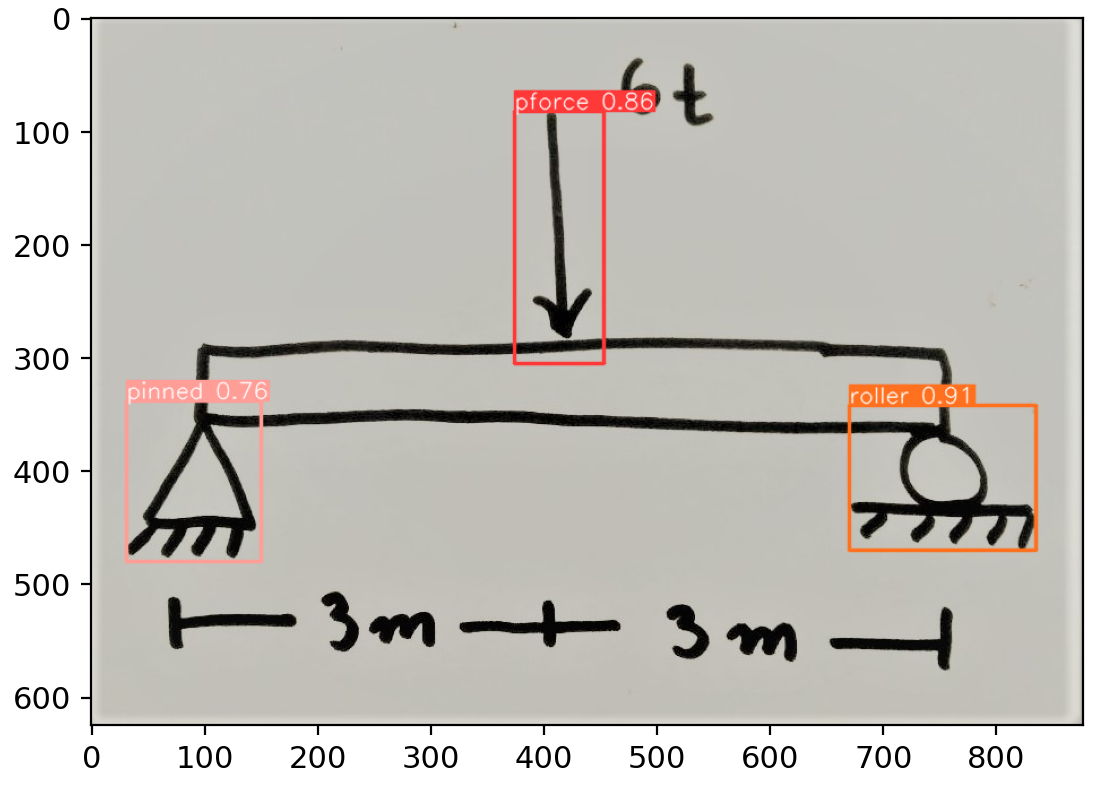}
		\label{yolo_usage}
	\end{subfigure}
	
	\caption{Object detection applied to supports of a theoretical beam (left) and an example of \textit{PhotoBeamSolver} using YOLOv5.(right)}
	\label{fig:combined}
\end{figure}

\begin{figure}[h!]
	\centering
	\includegraphics[width=0.8\linewidth]{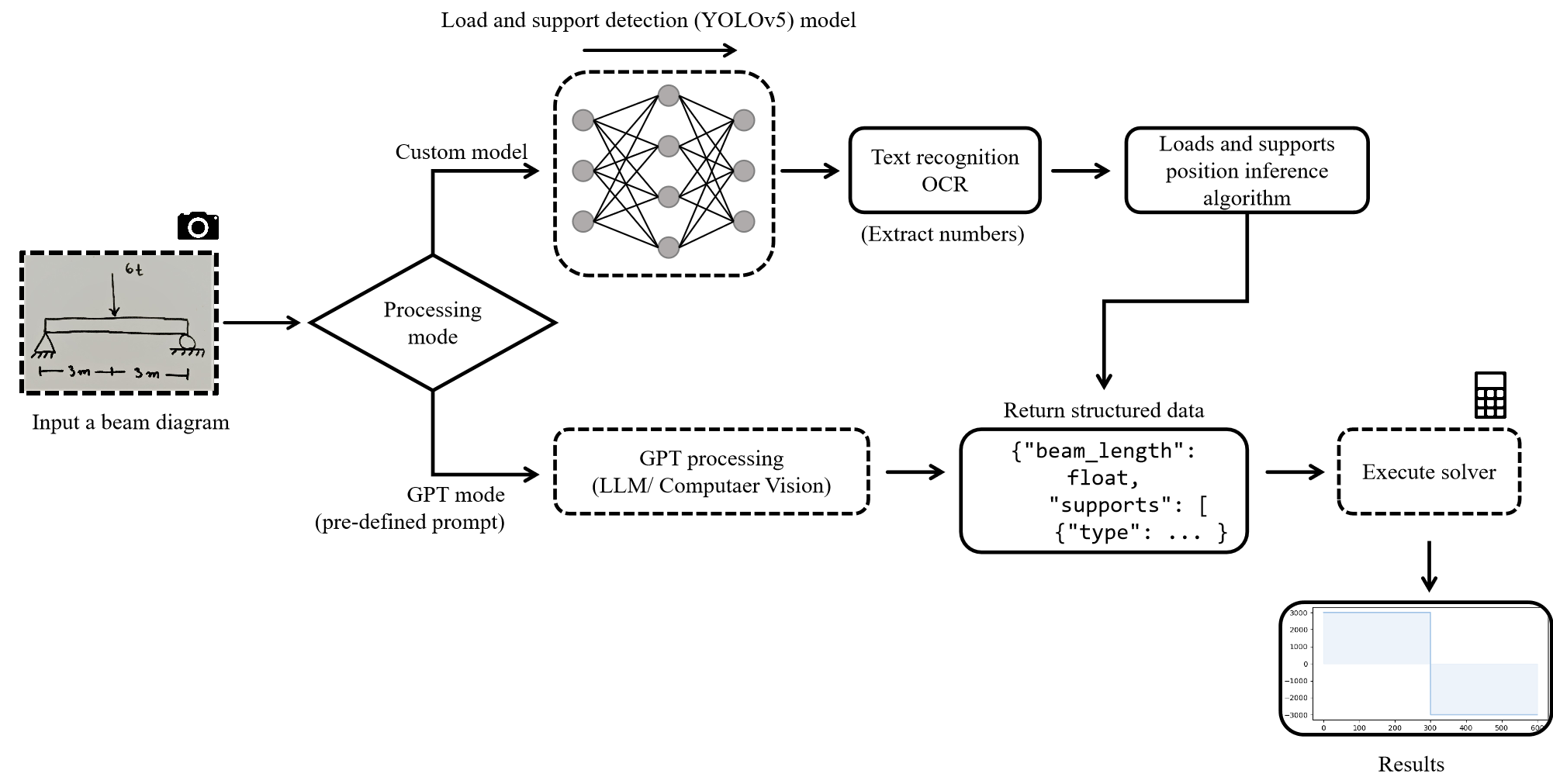}
	\caption{Flow chart of PhotoBeamSolver}
	\label{scheme}
\end{figure}

The training of the deep learning model $f$ is formulated within a supervised learning framework \cite{found_mohri}, where labeled pairs $(\mathbf{x}_i, y_i)$ are used to approximate the functional mapping between visual input and structural interpretation. In this context, the objective is to identify an optimal hypothesis $h^*: h^* \in \mathcal{H}$ that generalizes to previously unseen beam diagrams. From a statistical standpoint, training corresponds to maximum likelihood estimation, or equivalently, to the minimization of an appropriate loss function, commonly cross-entropy in detection tasks.\\
Given the problem addressed in this work, automatic interpretation of hand-drawn idealized beam models, the dataset $\mathcal{D}$ must capture the variability in drawing styles, geometric configurations, support types, and loading conditions typically encountered in academic structural analysis. Representativeness is critical, as the deployed system must infer structural topology and boundary conditions from imperfect visual inputs. The dataset is partitioned into a training subset $\mathcal{D}_t$ and an evaluation subset $\mathcal{D}_c$, the latter reserved exclusively for performance assessment. This separation ensures an unbiased estimate of generalization capability.\\
In the specific case of object detection applied to structural components, such as supports, loads, and beam segments, the learning task extends beyond classification. The model must localize elements within the image and preserve spatial relationships, as structural equilibrium and internal force computation depend directly on relative positioning. Consequently, the loss formulation integrates both localization and classification terms, aligning probabilistic prediction with geometric consistency. Notably, relative positioning is not used directly for structural computation. Instead, it serves as an intermediate geometric reference for the subsequent positioning inference algorithm, which identifies and interprets handwritten annotations such as span lengths and support locations.\\
Once the algorithm is selected, the model is trained using the training subset $\mathcal{D}_t$. This stage consists of estimating the parameter tensor $\textbf{w}$ such that the predicted output approximates the ground truth for each input sample. Formally, the parameters are adjusted to minimize a predefined loss function over the training data.\\
After convergence, the trained model is evaluated on the test subset $\mathcal{D}_c$ in order to quantify its predictive performance. This evaluation is performed by comparing the predicted outputs (e.g., $y_k$ in classification or $y_i$ in regression) with their corresponding ground-truth values. The resulting metrics provide an estimate of the model’s generalization capability.
\begin{figure}[!htb]
	\minipage{0.32\linewidth}
	\includegraphics[width=\linewidth]{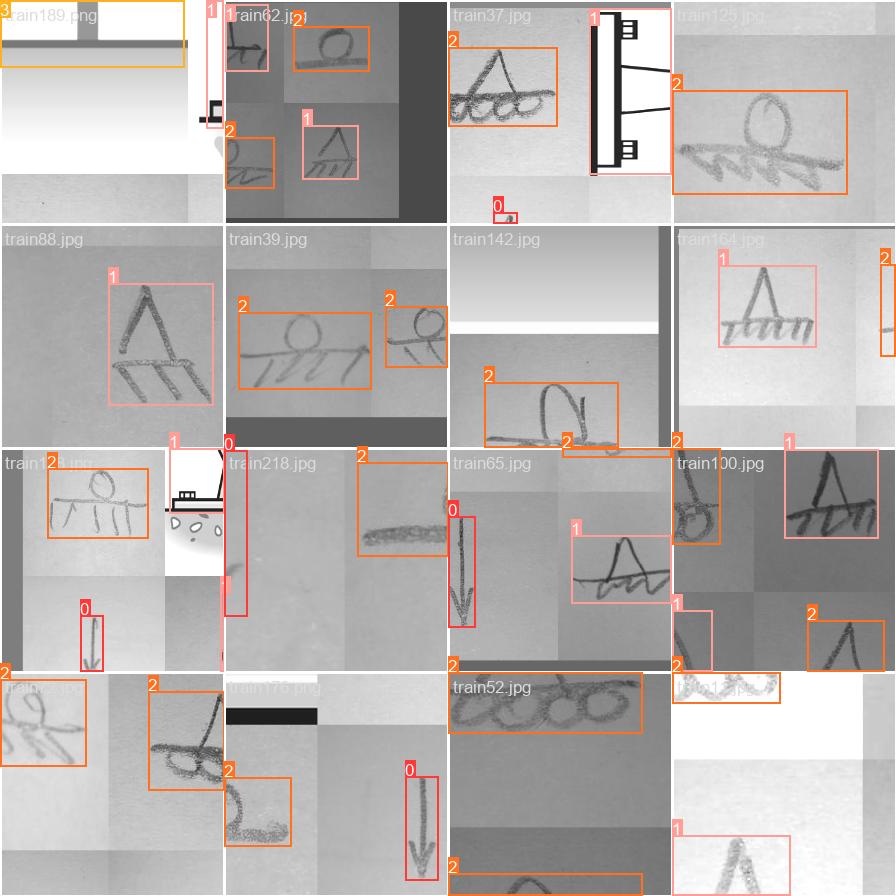}
	\endminipage\hfill
	\minipage{0.32\linewidth}
	\includegraphics[width=\linewidth]{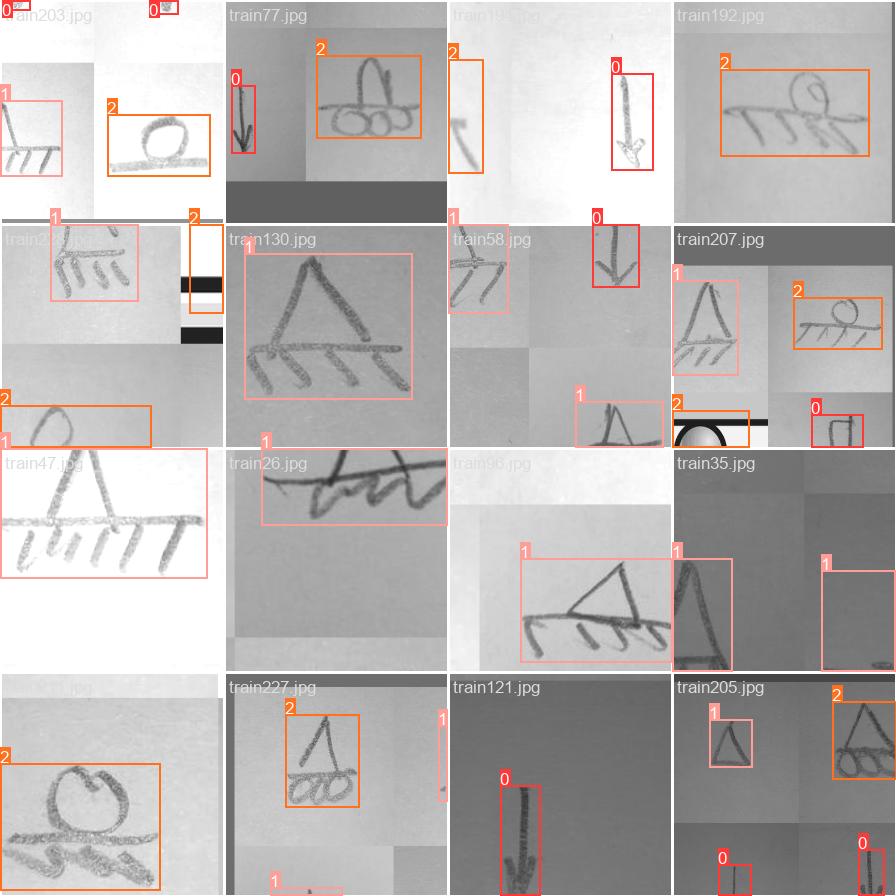}
	\endminipage\hfill
	\minipage{0.32\linewidth}%
	\includegraphics[width=\linewidth]{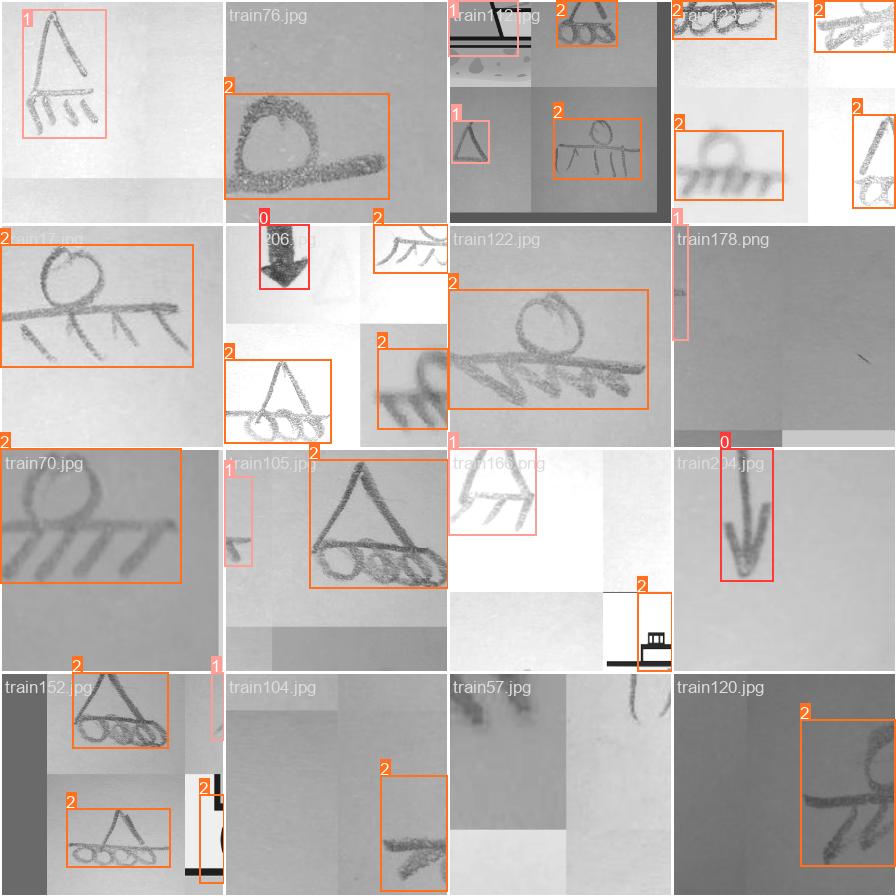}
	\endminipage
	\caption{Examples of training datasets}
	\label{train_datasets}
\end{figure}
The output of the detection model consists of spatially localized structural components expressed in image coordinates. These detections are transformed into a structured mechanical representation through geometric post-processing.\\
Relative positions of supports and loads are computed from normalized bounding box coordinates. Beam span length is inferred from endpoint detections or reference markers. Load positions are projected onto the beam axis, and support types are encoded according to their detected classes. When magnitude annotations are present in the diagram, optical or positional cues are used to associate loads with their corresponding values.\\
This stage produces a symbolic description of the structural system, including boundary conditions, load types, load positions, and geometric parameters. The representation is serialized into a structured format suitable for downstream analytical processing. Importantly, this intermediate abstraction decouples image interpretation from structural reasoning, reducing ambiguity before symbolic computation is attempted, as shown in the flowchart of figure \ref{scheme}.

\subsubsection*{LLM-Based Analytical Reasoning}
Once the structured representation module is generated, it is passed to an analytical solver module. This module can operate using either a custom-trained object detection or a Large Language Model (LLM) outputs, depending on the selected configuration. When an LLM is employed, it operates exclusively on structured textual instructions and image-derived input produced during the detection stage.\\
The solver module is responsible for interpreting boundary conditions, formulating equilibrium equations, and deriving reaction forces and internal force expressions.\\
The use of LLMs, when enabled, functions as a parameters detector and interpreter rather than as a numerical regression model. Instead of approximating structural response through data-driven prediction, the system produces analytical formulations consistent with classical structural mechanics. This design preserves flexibility in problem formulation.\\
A strict separation between visual detection and solver is maintained. By constraining the solver module to structured inputs, the system limits the propagation of visual ambiguities and, in the case of LLM deployment, mitigates hallucination risks by grounding analytical outputs in explicitly detected structural parameters.

\begin{figure}
	\includegraphics[width=\linewidth]{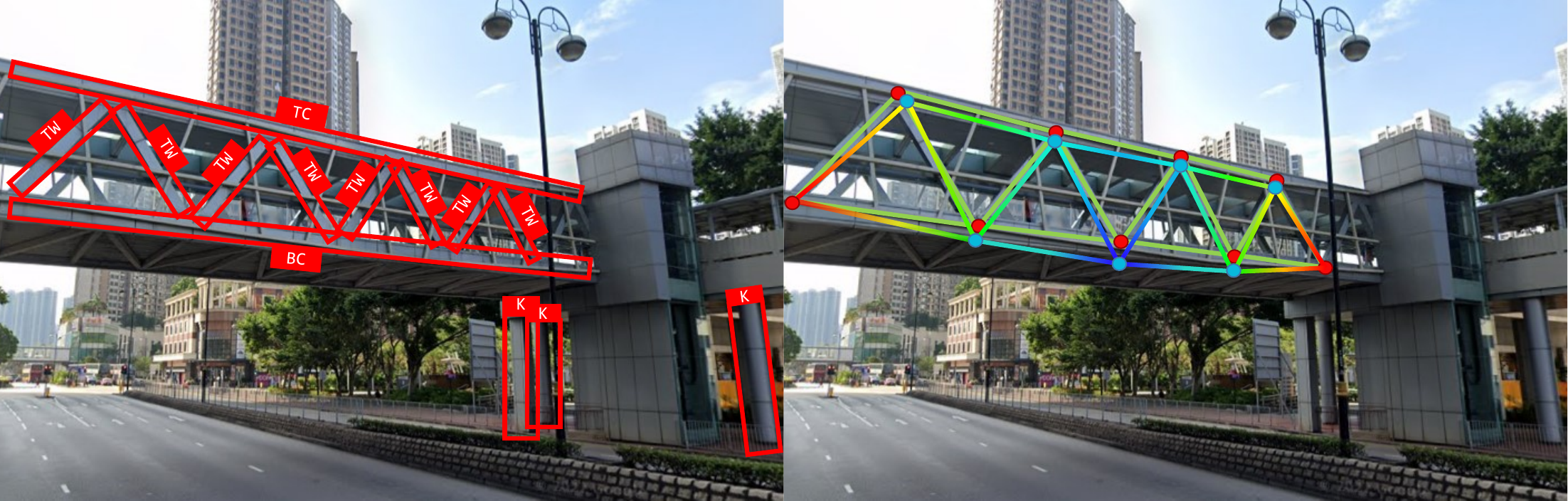}
	\caption{Conceptualization of a computer vision based structural analysis.}
	\label{real_truss_detec}
\end{figure}
\section{Discussion}
\label{cap:3}
In the context of computer-based structural analysis, which infers structural responses directly from beam diagrams, it becomes necessary to quantify the reliability of the predictions, leading to the problem of uncertainty quantification, which refers to the estimation of the uncertainty associated with the prediction produced by a statistical learning model. This form of uncertainty is commonly termed epistemic uncertainty \cite{H_llermeier_2021}, as it arises from incomplete knowledge of the underlying data-generating process, often referred to as the ground truth. A standard deep learning model implements a deterministic mapping from an input $\mathbf{x}_i$ to an output $y_i$, without explicitly characterizing the confidence or dispersion of that prediction. Even when the final layer is a \textit{softmax} function, the output represents normalized scores rather than a calibrated measure of predictive certainty, and does not provide a statistically grounded confidence interval \cite{pml2Book}.\\
In addition to epistemic uncertainty, random uncertainty may also be present; however, the latter is intrinsic to data variability and is fundamentally probabilistic in nature.\\
Quantifying predictive uncertainty becomes necessary when model outputs are used to support decision-making processes. In the specific case of \textit{PhotoBeamSolver}, moderate predictive uncertainty is not critical in controlled usage scenarios, particularly when a human operator (professional) verifies and, if necessary, corrects detected structural elements. However, in more demanding contexts—such as automated interpretation of real structural systems as illustrated in figure \ref{real_truss_detec}, both predictive accuracy and reliability become essential, making uncertainty estimation a relevant component of the framework.\\
A common misconception is that \textit{softmax} outputs provide a direct measure of predictive confidence. For instance, object detection frameworks such as YOLOv5 (Ultralytics) report a quantity termed “confidence”, which is derived from these normalized outputs. Nevertheless, \textit{softmax} probabilities do not necessarily reflect calibrated predictive certainty. Consider a poorly trained binary classifier whose final layer is \textit{softmax}. Even under inadequate optimization or low-quality data, the maximum class probability cannot fall below 0.5. Consequently, for a new input $\mathbf{x}_{n+1}$, the model may assign a probability exceeding 0.5 to an incorrect class, despite lacking meaningful predictive reliability. For this reason, \textit{softmax} outputs are generally regarded as heuristic indicators rather than rigorous measures of uncertainty \cite{a_gentle_intro_conf}.\\
That is the case for custom-trained models, which is at least conceptually grounded in statistical learning theory. In contrast, quantifying uncertainty in LLMs symbolic reasoning is substantially more challenging, as token-level probabilities do not directly correspond to confidence in derived analytical formulations. This distinction underscores the importance of constraining LLM outputs to structured, verifiable representations when applied to engineering problems.
\section{Results}
\label{cap:4}
\begin{figure}[!t]
	\centering
	\includegraphics[width=0.6\linewidth]{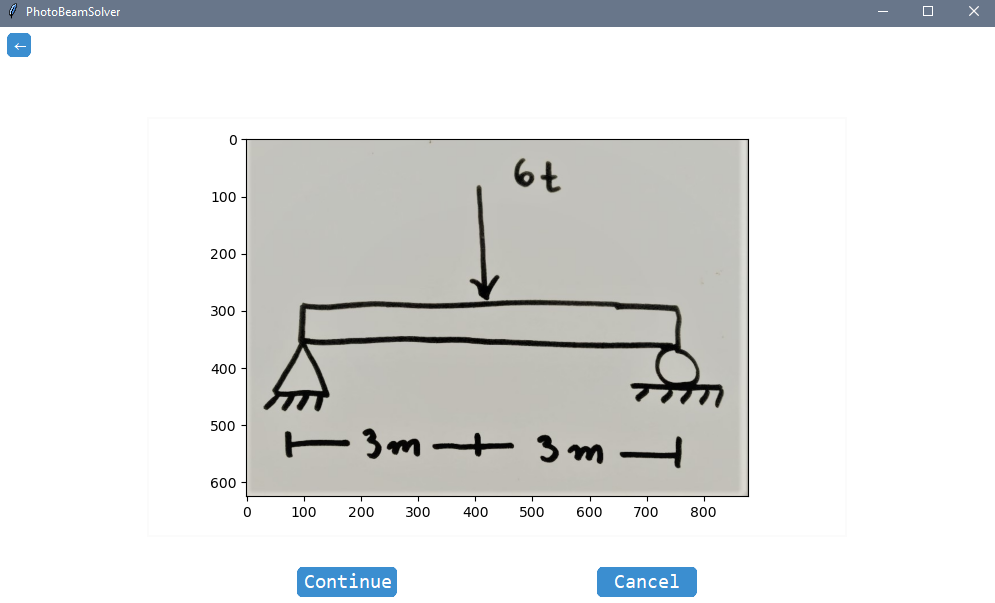}
	\caption{Input beam}
	\label{input}
\end{figure}

\begin{figure}[!h]
	\centering
	
	\begin{subfigure}[t]{0.32\linewidth}
		\centering
		\includegraphics[width=\linewidth]{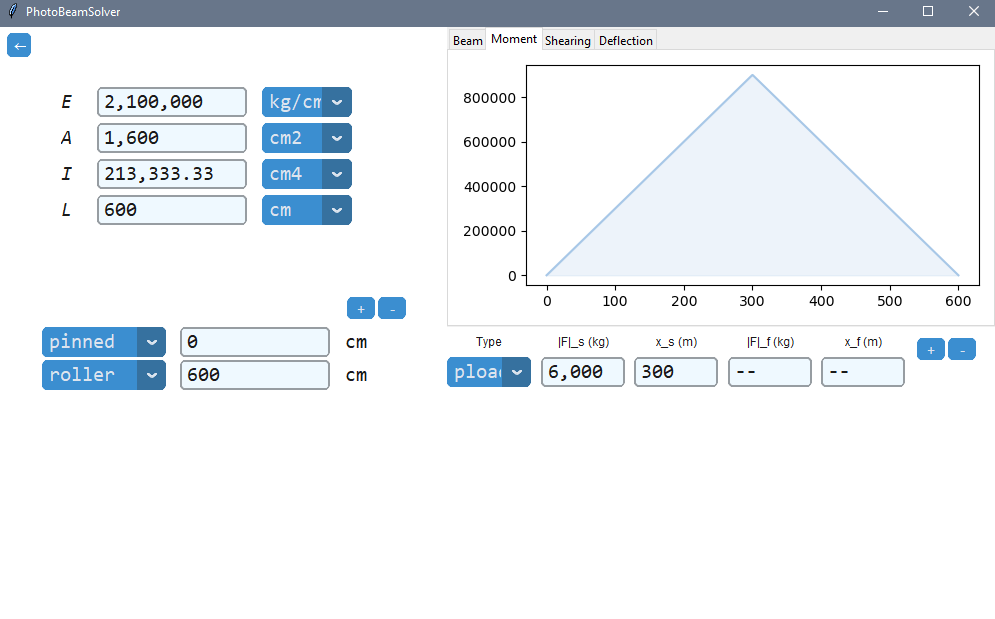}
	\end{subfigure}
	\hfill
	\begin{subfigure}[t]{0.32\linewidth}
		\centering
		\includegraphics[width=\linewidth]{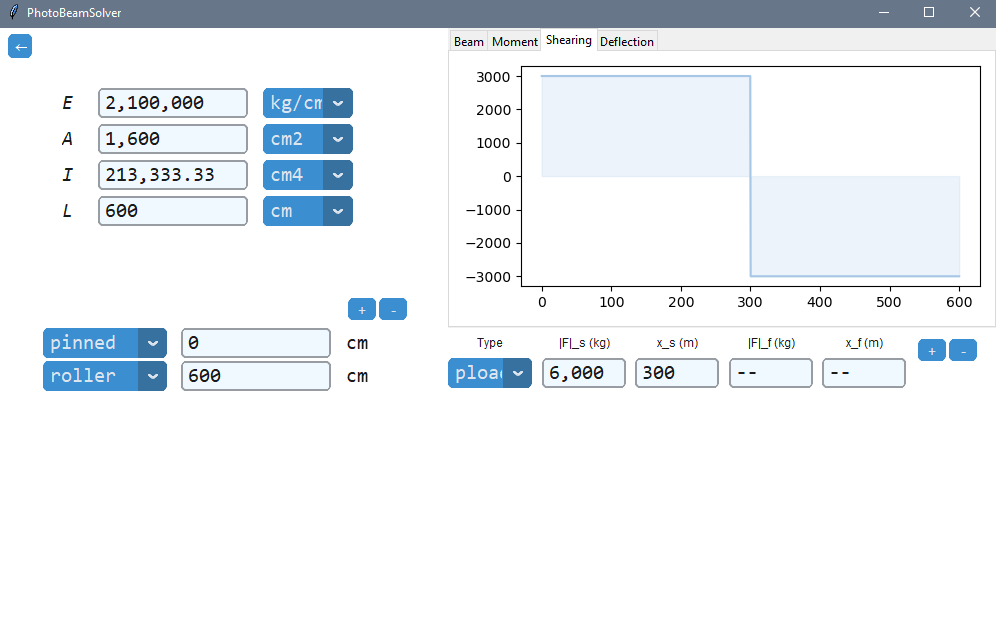}
	\end{subfigure}
	\hfill
	\begin{subfigure}[t]{0.32\linewidth}
		\centering
		\includegraphics[width=\linewidth]{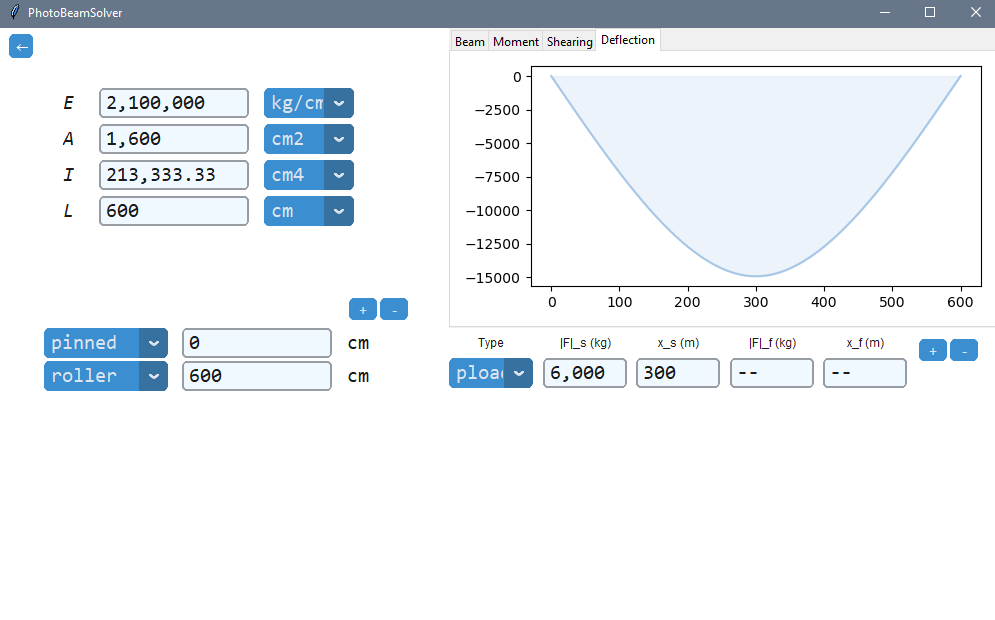}
	\end{subfigure}
	
	\caption{Bending moment diagram displayed in PBS. Units [kg·cm] (left), shear force diagram displayed in PBS. Units [kg] (center) and deflection diagram displayed in PBS. Units [mm] (right)}
	\label{pbs_diagrams}
\end{figure}

The training results are presented in Table \ref{cuadro_resultados}. While the beam response can be determined by detecting actions (forces and moments) and geometric parameters (distances and supports), the program \textit{PhotoBeamSolver} also incorporates beam deflection analysis, which requires additional parameters such as Young’s modulus and second moment of area.\\
\textit{PhotoBeamSolver} uses the GPT API when a valid API session is available in the system; otherwise, it relies on the custom-trained detection model to process the input image.\\
The object detection model was trained on a dataset of 532 annotated beam diagrams. The dataset includes supports (pinned, roller, fixed), point loads, distributed loads, and reference geometric markers. Training was performed under a supervised learning framework, and performance was evaluated using mean Average Precision (mAP) and per-class precision metrics.\\
Convergence behavior indicates rapid performance improvement during early epochs, followed by gradual stabilization. Final mAP values above 0.93 demonstrate high localization and classification accuracy for the defined structural components. Per-class precision remains consistently high across support types and load categories, indicating balanced learning without significant class bias.\\
To assess generalization capability, evaluation was conducted on a held-out validation subset. The stability of mAP and per-class precision across late epochs suggests that the model does not exhibit significant performance degradation. However, given the relatively limited dataset size, the system remains constrained to idealized diagram styles similar to the training distribution.\\
After detection, geometric post-processing converts bounding box outputs into a structured mechanical representation. Support positions are projected onto the beam axis, load locations are normalized with respect to beam span, and magnitudes are associated through positional proximity to annotations.\\
The structured representation is then processed by the solver module. The custom-trained detection and deterministic processing pipeline is used exclusively. In both configurations, the solver produces reaction forces, shear force expressions, bending moment distributions, and deflection curves.\\
Analytical consistency was verified by comparing computed reactions and internal force diagrams against reference solutions obtained using the IndeterminateBeam Python package. Numerical discrepancies remained within machine precision for statically determinate cases, confirming the correctness of the symbolic formulation and integration steps.\\
Figure \ref{input} illustrates a representative input diagram. Figure \ref{pbs_diagrams} show the resulting shear force, bending moment, and deflection diagrams, respectively. The system successfully reconstructs classical structural response curves consistent with beam theory.\\
Failure cases primarily arise from ambiguous handwriting, overlapping annotations, or incomplete geometric references. In such scenarios, detection confidence decreases and structured interpretation may require manual correction.\\
Overall, the results validate the feasibility of the proposed image-to-analysis pipeline for idealized planar beam configurations.

\begin{table}[h!]
	\centering
	\small
	\begin{tabular}{lllllllll}
		\hline
		Epoch  & Precision & mAP       & \multicolumn{6}{c}{Error}                                                 \\
		\hline
		&           &           & pload                 & dload     & fixed      & roller     & simple      &   moment   \\
		\hline
		1      & 0.012942  & 0.0089532 & 0.073366              & 0.0083674 & 0.036995   & 0.00920414 & 0.03551520 & 0.036138 \\
		2      & 0.66248   & 0.022613  & 0.068911              & 0.0088902 & 0.039631   & 0.00977922 & 0.03804576 & 0.038322 \\
		3      & 0.56978   & 0.017702  & 0.063276              & 0.0078639 & 0.033436   & 0.00865029 & 0.03209856 & 0.032504 \\
		4      & 0.67439   & 0.059428  & 0.057429              & 0.0081983 & 0.029804   & 0.00901813 & 0.02861184 & 0.028587 \\
		5      & 0.63468   & 0.12734   & 0.056769              & 0.0082484 & 0.024758   & 0.00907324 & 0.02376768 & 0.024040 \\
		6      & 0.54589   & 0.13552   & 0.051374              & 0.0067728 & 0.020269   & 0.00745008 & 0.01945824 & 0.019040 \\
		7      & 0.69743   & 0.24772   & 0.041172              & 0.0066267 & 0.016964   & 0.00728937 & 0.01628544 & 0.015362 \\
		8      & 0.68126   & 0.30173   & 0.047487              & 0.0061217 & 0.016579   & 0.00673387 & 0.01591584 & 0.016207 \\
		9      & 0.58848   & 0.29588   & 0.041639              & 0.0058496 & 0.013002   & 0.00643456 & 0.01248192 & 0.013072 \\
		10     & 0.78813   & 0.34182   & 0.037008              & 0.005601  & 0.011118   & 0.0061611  & 0.01067328 & 0.010019 \\
		&           &           &  $\vdots$ &           &            &            &             \\
		9,998  & 0.99651   & 0.94911   & 0.0063394             & 0.0013403 & 0.00031973 & 0.00147433 & 0.000306941 & 0.000657 \\
		9,999  & 0.99735   & 0.95319   & 0.0062695             & 0.0013347 & 0.00031992 & 0.00146817 & 0.000307123 & 0.000292 \\
		10,000 & 0.99711   & 0.93829   & 0.0066504             & 0.0013822 & 0.00033773 & 0.00152042 & 0.000324221 & 0.000055 \\
		\hline
	\end{tabular}
	\caption{Training results of the neural network using a dataset of 532 images}
	\label{cuadro_resultados}
\end{table}

Here, \textit{pload} refers to point load data, \textit{dload} to distributed load data, \textit{roller} to roller supports, \textit{fixed} to fixed (clamped) supports, and \textit{simple} to pinned supports.

\section{Conclusions}
\label{cap:conc}
The system does not rely on fully connected neural networks for image interpretation, as such architectures fail to preserve spatial locality and exhibit poor scalability when processing high-dimensional image inputs. Convolutional architectures provide translation robustness and parameter efficiency, which are essential for consistent detection of structural components regardless of diagram placement or scale.\\
Object detection is employed instead of pure image classification because structural analysis depends not only on the presence of components but also on their spatial configuration. Reaction forces and internal force distributions are functions of relative positioning; therefore, accurate localization is necessary to construct valid equilibrium equations. A classification-only approach would be insufficient to encode the geometric relationships required for deterministic structural analysis.\\
The integration of Large Language Models allows the system dynamically generates structured analytical derivations from the detected symbolic representation, eliminating the need for custom trained models but it requires the API keys of the corresponding LLM.\\
The results demonstrate that \textit{PhotoBeamSolver} is capable of successfully detecting idealized beam models and resolving them analytically, producing consistent shear force, bending moment, and deflection diagrams. The combined perception pipeline performs reliably for the class of problems considered, validating the feasibility of integrating computer vision with LLM-based analysis for structural applications.\\
However, the applicability of the system remains constrained. The current implementation does not account for inclined loads, truss systems, or inclined beams. Inclined loads must be decomposed into orthogonal components prior to processing, and inclined beams are treated within their local coordinate systems as horizontally equivalent members. These simplifications limit the system to planar, predominantly statically determinate beam configurations.\\
Extending the approach towards more general structural analysis would introduce significant uncertainty. Parameters such as \textbf{design loads, material properties, cross-sectional characteristics}, and \textbf{construction details} cannot be reliably inferred from a single image. Although design codes provide normative frameworks, estimating such parameters from visual data alone would require a probabilistic treatment of uncertainty that lies beyond the scope of this work but may warrant consideration in future research.\\
Another major obstacle to scaling this approach is data availability. Unlike conventional computer vision tasks, suitable training data cannot be generated through generic labeling services. Structural datasets require well-documented engineering models with associated design metadata provided by practitioners. The acquisition of such data is nontrivial and represents a limiting factor for broader generalization.\\
Despite these constraints, \textit{PhotoBeamSolver} demonstrates practical utility, particularly in academic contexts and in specific professional scenarios involving idealized beam analysis. While currently implemented as a desktop application, the trained neural network can be integrated into web-based beam calculators or mobile applications as an \textbf{auxiliary feature}, expanding accessibility without altering the underlying methodology.\\
Overall, the proposed framework establishes a viable image-to-analysis pipeline by combining domain-specific object detection with LLM-driven reasoning and approximate structural solvers. The contribution lies in demonstrating that structural interpretation and analytical derivation can be coherently unified within a computer vision assisted system.
\section*{Programs}
The script required to execute the \textit{PhotoBeamSolver} implementation is available in the following GitHub \href{https://github.com/xiaixue/pbs}{repository}.

\phantomsection
\addcontentsline{toc}{section}{Referencias}
\renewcommand\bibname{Referencias}
\setstretch{1.1}
\bibliographystyle{unsrt}
\bibliography{references}

\appendix
\end{document}